%% file: main.tex
\documentclass[10pt,twocolumn,letterpaper]{article}

\usepackage[pagenumbers]{cvpr} 

\usepackage{enumitem}
\usepackage{graphicx}
\usepackage{mathabx}
\usepackage{pifont}

\usepackage{svg}

\usepackage{float}

\usepackage{subcaption}
\usepackage{multirow, makecell}

\usepackage{algorithm}
\usepackage{algpseudocode}  
\usepackage{amssymb}
\usepackage{amsmath}  
\usepackage{amsthm}
\allowdisplaybreaks[4]


\definecolor{cvprblue}{rgb}{0.21,0.49,0.74}
\usepackage[pagebackref,breaklinks,colorlinks,citecolor=cvprblue]{hyperref}

\def\paperID{} 
\def\confName{CVPR}
\def\confYear{2024}

\title{On the Road to Portability: Compressing End-to-End Motion Planner for Autonomous Driving}

\author{Kaituo Feng$^1$, \quad Changsheng Li$^{1}$\thanks{Corresponding author}, \quad 
Dongchun Ren$^{2}$, \quad Ye Yuan$^{1}$, \quad Guoren Wang$^{1,3}$\\
\textsuperscript{1} Beijing Institute of Technology \quad 
\textsuperscript{2} ALLRIDE.AI \\
\textsuperscript{3} Hebei Province Key Laboratory of Big Data Science and Intelligent Technology \\
{\tt\small kaituofeng@gmail.com, lcs@bit.edu.cn, Dongchun.ren@allride.ai} \\ 
{\tt\small yuan-ye@bit.edu.cn, wanggrbit@126.com}
}

\begin{document}
\maketitle
\input{sec/0_abstract}    
\input{sec/1_intro}
\input{sec/2_related}
\input{sec/3_method}

\input{sec/4_experiment}
\input{sec/5_conclusion}

{
    \small
    \bibliographystyle{ieeenat_fullname}
    \bibliography{main}
}

\input{sec/X_suppl}


\end{document}

%% file: sec/0_abstract.tex
\begin{abstract}
End-to-end motion planning models equipped with deep neural networks have shown great potential for enabling full autonomous driving. However, the oversized neural networks render them impractical for deployment on resource-constrained systems, which unavoidably requires  more computational time and resources during reference.  
To handle this, knowledge distillation offers a promising approach that compresses models by enabling a smaller student model to learn from a larger teacher model. Nevertheless, how to apply knowledge distillation to compress motion planners has not been explored so far. In this paper, we propose PlanKD, the first knowledge distillation framework tailored for compressing end-to-end motion planners. First, considering that driving scenes are inherently complex, often containing planning-irrelevant or even noisy information, transferring such information is not beneficial for the student planner. Thus, we design an information bottleneck based strategy to only distill  planning-relevant information, rather than transfer all information indiscriminately. Second, different waypoints in an output planned trajectory may hold varying degrees of importance for motion planning, where a slight deviation in certain crucial waypoints might lead to a collision. Therefore, we devise a safety-aware waypoint-attentive distillation module that assigns adaptive weights to different waypoints based on the importance, to encourage the student to accurately mimic more crucial waypoints, thereby improving overall safety. Experiments demonstrate that our PlanKD can boost the performance of smaller planners by a large margin, and significantly reduce their reference time. 
\end{abstract}

%% file: sec/1_intro.tex
\vspace{-0.15in}
\section{Introduction}
\label{sec:intro}

\begin{figure}
  \begin{center}
    \includegraphics[width=0.45\textwidth]{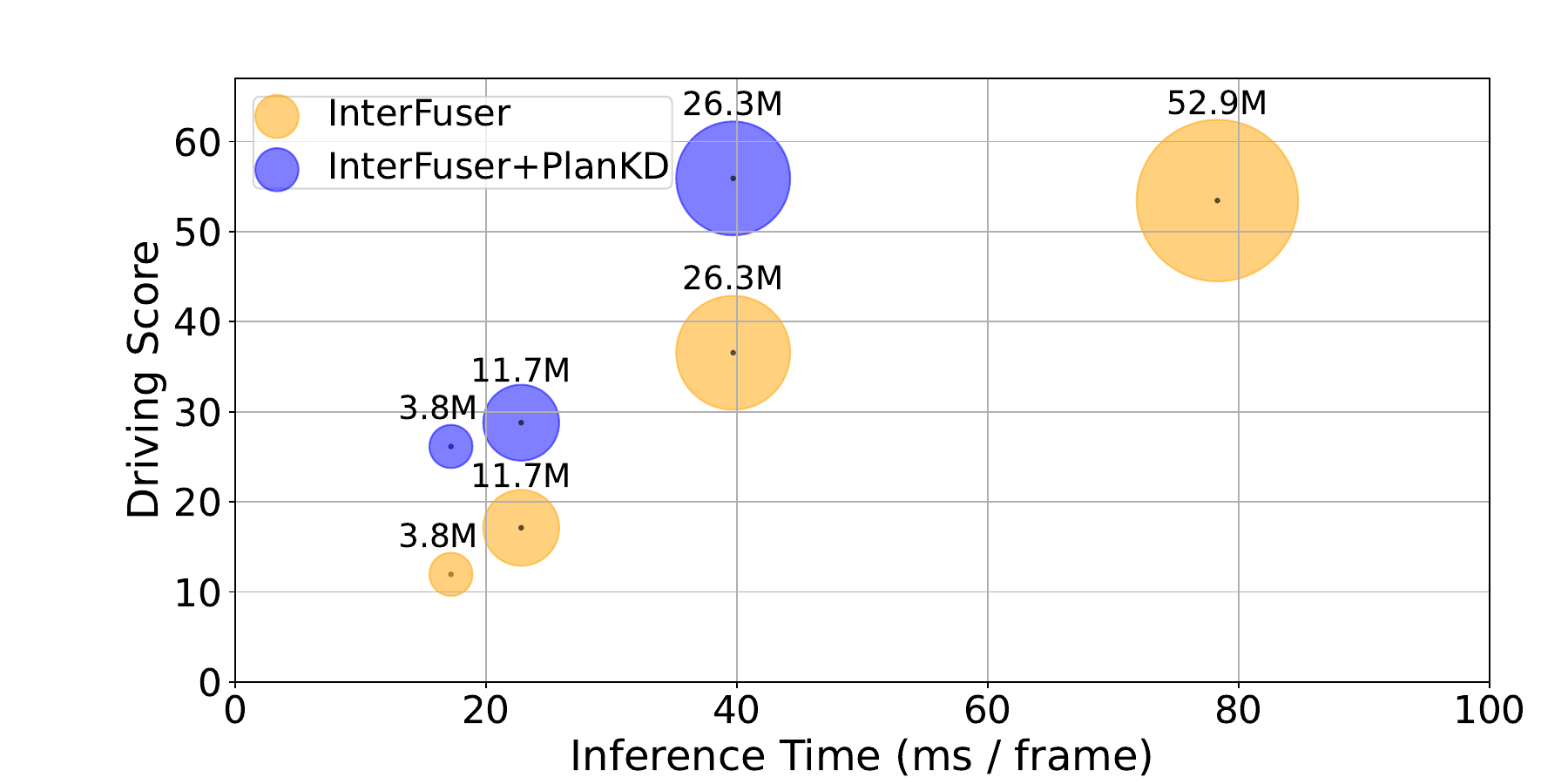}
  \end{center}
    \vspace{-0.2in}
  \caption{ An illustration for the performance degradation of InterFuser \cite{shao2023safety} on Town05 Long Benchmark \cite{prakash2021multi}  as the number of parameters decreases. By leveraging our PlanKD, the performance of compact motion planners can be enhanced, and the inference time can be significantly lowered. The inference time is evaluated on GeForce RTX 3090 GPU in a server. Best viewed in color.
  }
  \vspace{-0.25in}
    \label{teaser}
\end{figure}

End-to-end motion planning has recently emerged as a promising direction in autonomous driving \cite{wutrajectory,prakash2021multi,chitta2021neat,zhang2021end,zhang2021learning,casas2021mp3,ohn2020learning}, which directly maps raw sensor data to planned motions.
This learning-based paradigm shows the merit of reducing heavy reliance on hand-crafted rules  and mitigating the accumulation of errors within intricate cascading modules (typically, detection-tracking-prediction-planning) \cite{wutrajectory,zhang2021end}.
Despite the success, the oversized architecture of deep neural networks in the motion planner 
poses challenges for deployment in resource-constrained environments, such as an autonomous delivery robot that relies on computing power from an edge device.
Furthermore, even within regular vehicles, the computational resources on onboard devices are often limited \cite{silva2021resource}. 
Thus, directly deploying deep and large planners unavoidably requires more computational time and resources during reference, making it challenging to respond rapidly to potential dangers.
To mitigate this issue, a straightforward approach is to reduce the number of network parameters by using smaller backbones, while we observe that the performance of the end-to-end planning model will drop dramatically, as shown in Figure \ref{teaser}.
For example, although the inference time of InterFuser \cite{shao2023safety} (a typical end-to-end motion planner) is lowered when reducing the number of parameters from $52.9$M to $26.3$M, its driving score drops from $53.44$ to $36.55$.
Therefore, it's necessary to develop a suitable model compression method tailored for end-to-end motion planning.

To derive a portable motion planner, we resort to knowledge distillation \cite{hinton2015distilling} for compressing end-to-end motion planning models  in this paper.
Knowledge distillation (KD) has been widely studied for model compression in various tasks, such as object detection \cite{chen2017learning,kang2021instance}, semantic segmentation \cite{he2019knowledge,liu2019structured}, etc.
The underlying idea of these works is to train a condensed student model by inheriting  knowledge from a larger teacher model, and utilize the student model as a substitute for the teacher model during deployment.
While these studies have achieved significant success, directly applying them to end-to-end motion planning would result in sub-optimal outcomes. This stems from two emerging challenges inherent in the task of motion planning:
(i) The driving scenarios are inherently complex \cite{zhan2017spatially}, involving a diverse array of information including multiple dynamic and static objects, intricate background scenes, as well as multifaceted road and traffic information. 
However, not all of these information is beneficial to planning. 
Background buildings and distant vehicles, for example, are irrelevant or even noise to  planning \cite{wu2023policy}, while nearby vehicles and traffic lights have a deterministic impact.
Thus, it is crucial  to automatically distill only planning-relevant information from the teacher model, while previous KD methods can not accomplish it.
(ii) Different waypoints in an output planned trajectory often hold varying degrees of importance for
motion planning.
For example, when navigating a junction, the waypoints proximate to other vehicles within the trajectory may carry higher significance than other waypoints.
This is because at such points, the ego-vehicle needs to actively interact with other vehicles, and even a minor deviation could lead to a collision.
However, how to adaptively determine crucial waypoints and accurately mimic them is another significant challenge for previous KD methods.

To tackle the above two challenges, we propose the first \textbf{K}nowledge \textbf{D}istillation method tailored for compressing end-to-end motion \textbf{Plan}ner in autonomous driving, called \textbf{PlanKD}. 
 First, we present a strategy grounded in the information bottleneck principle \cite{alemi2016deep}, with the goal of distilling planning-relevant features that contains 
  the minimum yet sufficient amount of information for planning. 
 Specifically, we maximize the mutual information between the extracted planning-relevant features and the ground truth of our defined planning states, while minimizing the mutual information between the extracted features and the intermediate feature map. 
This strategy enables us to distill only the essential planning-relevant information in intermediate layers, thus enhancing the effectiveness of the student.
Second, to dynamically identify crucial waypoints and faithfully mimic them, we employ an attention mechanism \cite{vaswani2017attention} to calculate an attention weight between  each waypoint and its associated context in bird-eye-view (BEV) representation of the driving scene. 
To promote accurate emulation of safety-critical waypoints during distillation, we design a safety-aware ranking loss that encourages higher attention weights for waypoints 
in close proximity to moving obstacles.
Accordingly, the safety of the student planner can be significantly enhanced.
As an evidence shown in Figure \ref{teaser}, the driving scores of the student planners can be significantly improved by our PlanKD. In addition, our method can lower the reference time by approximately 50\%, while preserving comparable performance to the teacher planner on the Town05 Long Benchmark.


Our contributions can be summarized as: 
1) We constitutes the first attempt to explore a dedicated knowledge distillation method to compress end-to-end motion planners in autonomous driving. 
2) We propose a general and novel framework PlanKD, which enables the student planner to inherit the planning-relevant knowledge in the intermediate layer, as well as fostering the accurate matching of crucial waypoints for improving safety.
3) Experiments illustrate that our PlanKD can improve the performance of smaller planners by a large margin, thereby offering a more portable and efficient solution for resource-limited deployment.


%% file: sec/2_related.tex
\section{Related Works}
In this section, we introduce related works including end-to-end motion planning and knowledge distillation.

\textbf{End-to-end motion planning.}
End-to-end motion planning models for autonomous driving usually directly take as input raw sensor data and output the planned trajectory or low-level actions \cite{codevilla2018end,liang2018cirl, hu2022st, xu2017end, hu2023planningoriented, prakash2021multi}. 
This learning-based paradigm can eliminate the need for heavy hand-crafted rules and reduce the accumulative errors in a complicated cascading modular design \cite{wutrajectory, prakash2021multi}. Recent years have seen a surge of research on end-to-end motion planning \cite{chen2022learning,wutrajectory, cui2021lookout}. 
For example, NEAT \cite{chitta2021neat} enables efficient reasoning for the spatial and temporal semantic information in driving scenario for end-to-end trajectory planning.  
Roach \cite{zhang2021end} and LBC \cite{chen2020learning} train an end-to-end motion planner by imitating a privileged agent that can access to the ground-truth state.
The work in \cite{wang2021learning} learns an interpretable end-to-end motion planning model called IVMP also by imitating a privileged agent, which additionally takes as input the optical flow. 
TCP \cite{wutrajectory} explores to integrate the low-level actions and the planning trajectory to derive a better planning strategy.
InterFuser \cite{shao2023safety} is proposed to provide a both interpretable and safe planning trajectory. 
The success of these works can be attributed to the strong representation ability of deep neural networks used in their models. However, the large number of parameters in deep models makes them difficult to deploy in resource-limited environments.

\textbf{Knowledge distillation.}
Knowledge distillation (KD) aims to enable a compact student model to mimic the behavior of a larger teacher model, thereby inheriting the knowledge embedded within the teacher model. 
KD has been widely studied for model compression in a variety of domains, such as computer vision \cite{hong2022cross,chen2022bevdistill,jang2022glamd,yang2022focal,zheng2022localization}, natural language processing \cite{sun2019patient,fu2021lrc,liu2022multi} and data mining \cite{chen2020self,feng2022freekd,yang2021extract}. 
For instance, 
AT \cite{zagoruyko2016paying} derives a light student model by distilling the attention map rather than the feature itself from the teacher.
ReviewKD \cite{chen2021distilling} attempts to use the knowledge in multiple layers of the teacher to teach one layer of the student. 
DKD \cite{zhao2022decoupled} decouples the logit distillation into target class distillation and non-target class distillation and separately distills the knowledge from these two parts. 
DPK \cite{zongbetter} dynamically incorporates part of knowledge in the teacher during distillation, enabling the distillation process at an appropriate difficulty.
Despite the success of these works, directly utilizing them for motion planning may yield sub-optimal results.
How to design a knowledge distillation method tailored for compressing an end-to-end motion planner has not been explored.

%% file: sec/3_method.tex
\section{Proposed Method}

\subsection{Preliminaries}
An end-to-end motion planner aims to produce a sequence of planned motions, enabling the ego-vehicle to arrive at a predetermined destination in time \cite{ohn2020learning}. Motion planners usually take as input state $I=\{o,m,c\}$ consisting of observation $o$, measurement $m$ and high-level command $c$. The observation $o$ denotes received sensor data, e.g., camera data or LiDAR data. The measurement $m$ is usually a current speed of ego-vehicle. The high-level command $c$ is a navigation signal usually consisting of $\{left, right, straight, follow\}$.
The output of a motion planner could be a planned trajectory, and can be sent into a PID Controller \cite{farag2020complex} to produce low-level control actions. Besides trajectory-based output, the motion planner could also directly output low-level control actions.
Since using trajectory-based output has shown the merit of accounting for a longer future horizon \cite{shao2023safety,wutrajectory},  
we focus on the trajectory-based output in this paper.
To train a motion planner, a popular method is imitation learning \cite{codevilla2018end}, which can be formulated as:
\begin{equation}
{\arg\min}_{\theta}\ \mathbb{E}_{(I,\mathcal{T}^*)\sim\mathcal{D}} [\mathcal{L}(\mathcal{F}_{\theta}(I),\mathcal{T}^*)],
\end{equation}
where $\mathcal{D}=\{(I,\mathcal{T}^*)\}$ denotes the dataset, containing state $I$ and the corresponding expert planned trajectory $\mathcal{T}^*=\{w^*_i\}^T_{i=1}$. $w^*_i$ is the $i^{th}$ waypoint of the planned trajectory and $T$ is the number of waypoints.
$\mathcal{F}_{\theta}$ is the motion planner with parameters $\theta$. The imitation loss function $\mathcal{L}$ usually adopts the absolute error (i.e. $L_1$ loss) between  waypoints output by $\mathcal{F}_{\theta}(I)$ and $\mathcal{T}^*$. In this way, the planner could learn to generate  a good planning trajectory by closely imitating the expert.

To achieve marvelous performance, the motion planner typically requires a significant number of parameters $\theta$, which hinders its deployment in resource-limited environment. Thus, we explore to employ the knowledge distillation technique to compress models for the motion planning task. We denote the teacher planner as $\mathcal{F}^T_{\theta}$ with parameters $\theta$ and the student planner as $\mathcal{F}_{\phi}^S$ with parameters $\phi$. Note that the number of parameters $\phi$ in the student model is far less than that in the teacher model $\theta$.   
Our target is to distill essential knowledge from  $\mathcal{F}^T_{\theta}$ to  $\mathcal{F}_{\phi}^S$,  so as to facilitate the use of the student model as a substitute for the teacher model during deployment.

\subsection{Framework Overview}
An end-to-end motion planner can be generally divided into two parts: the perception backbone and the motion producer \cite{tampuu2020survey}. The former is responsible for understanding the driving scene and encoding the environment information, e.g., ResNet \cite{he2016deep} is often used as the backbone  \cite{wutrajectory, chen2022learning, shao2023safety}. The latter receives the encoded information and generates a planned trajectory  \cite{wutrajectory,shao2023safety}.
To enable the compact student planner to inherit knowledge from the larger teacher planner, we attempt to distill knowledge from both the intermediate feature maps in the perception module and the output planned waypoints in the motion producer module.

\begin{figure*}
  \centering
  \includegraphics[width=0.86\linewidth]{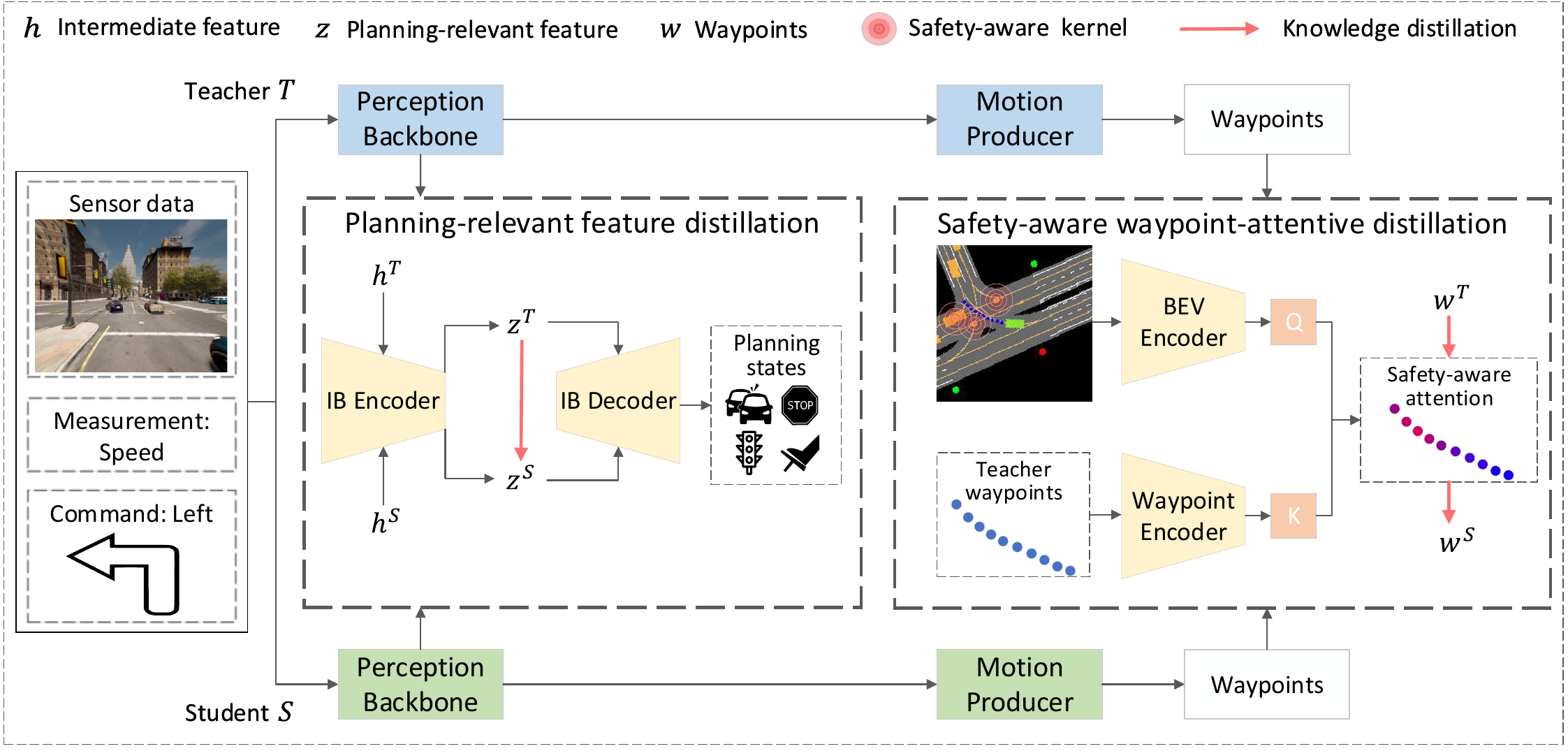}
    \vspace{-0.1in}
  \caption{
    An illustration of our PlanKD framework. PlanKD consists of two modules: a planning-relevant feature distillation module distilling planning-relevant features from intermediate feature maps via information bottleneck (IB); a safety-aware waypoint-attentive distillation module that dynamically determines crucial waypoints and distills knowledge from them for overall safety.
  }
  \label{main}
  \vspace{-0.2in}
\end{figure*}

As aforementioned, there are two key issues for distilling knowledge to a compact student planner: 
1) The knowledge encoded in intermediate feature maps could contain numerous planning-irrelevant or even noisy information. How to filter out such information is the key to knowledge distillation in a motion planning task. 
2) In an output planned trajectory, each waypoint may hold different levels of importance. A knowledge distillation method should possess the capability to learn and transfer this information to ensure safety.
Thus, we propose PlanKD, of which the framework is shown in Figure \ref{main}. 
Our PlanKD consists of two main modules. Firstly, we devise a planning-relevant feature distillation module. It utilizes the information bottleneck principle to extract planning-relevant information from intermediate feature maps, and transfer this information to the student model for effective distillation. 
Moreover, we also design a safety-aware waypoint-attentive distillation module. This module can assign  adaptive weights for waypoints in a trajectory based on their importance, and distill such information for improving overall safety. Next, we will elaborate the two modules in our PlanKD.

\subsection{Planning-relevant Feature Distillation}
Considering that driving scenes usually contain numerous planning-irrelevant or even noisy information, we intend to leverage the information bottleneck principle to only transfer the planning-relevant information during distillation. 

\textbf{Learning planning-relevant feature via information bottleneck.}
The core concept of the information bottleneck is to learn a representation that simultaneously minimizes the correlation between  the representation and inputs while maximizing the correlation between the representation and the class \cite{tishby2000information}.
In this paper, we attempt to extend information bottleneck to learn planning-relevant features.
Specifically, we intend to derive a planning-relevant representation by minimizing the mutual information between the representation and the intermediate feature map while maximizing the mutual information between the representation and the ground truth of the planning states (to be introduced later). Our objective could be formulated as:
\begin{equation}\label{ibobj}
\setlength{\abovedisplayskip}{0.05in}
\setlength{\belowdisplayskip}{0.05in}
    J_{IB}=\max_{Z}\ \sum_{i=1}^M I(Z,Y^i)-\beta I(Z,H),
\end{equation}
where $\beta$ is the Lagrange multiplier.
$I(\cdot,\cdot)$ denotes the mutual information and $M$ is the number of planning states. $Z$ is the learned planning-relevant representation. $H$ and $Y^i$ are random variables of the intermediate feature map and ground truth of $i^{th}$ planning state, respectively. Before introducing how to distill knowledge from $Z$, we first define the planning states used in this paper.

\textbf{Planning states.}
The planning states summarize some essential aspects of a motion planning task.
We define two kinds of planning states: environment states and action states. The environment states are used to encapsulate the status of some elements that are influential or deterministic to planning in the environment. Moreover,  we introduce actions states to provide a summary of the ego-vehicle's current motion status. The action states could indicate whether the ego-vehicle encounters with some situations that requires take these actions. 

To be specific, we define eight planning states, consisting of five environment states and three low-level action states.
The environment states includes: nearby vehicle state, nearby pedestrian state, traffic sign state, junction state and traffic light state. 
The first four environment states are represented as binary indicators, signifying the presence or absence of these elements in the surrounding environment. The traffic light state adopts a three-value representation, denoting its absence, red, or green state.
Furthermore, the low-level action states encompass brake state, throttle state and steer state. These binary states record whether the magnitude of these actions exceeds a certain threshold $\delta$.

\textbf{Planning-relevant feature distillation.}
After defining the planning states $Y^i$, we can learn the planning-relevant representation $Z$ by maximizing $\sum_{i=1}^M I(Z,Y^i)$ while minimizing $\beta I(Z,H)$, as in Eq.(\ref{ibobj}). Note that  it's intractable to directly optimize Eq.(\ref{ibobj}), thus we utilize the method in  \cite{alemi2016deep} to estimate its 
lower bound:
\begin{align} \nonumber
\setlength{\abovedisplayskip}{0.05in}
\setlength{\belowdisplayskip}{0.05in}
     J_{IB} \geq \mathcal{L}_{IB} = & \frac{1}{N} \sum_{i=1}^N 
  \{\mathbb{E}_{\epsilon} [ \sum_{j=1}^M \log q_d(y_i^j|f_e(h_i,\epsilon)) ]\} \\ 
   & -  \frac{\beta}{N} \sum_{i=1}^N  \mathrm{KL} [p(Z|h_i)||r(Z)], 
\end{align} 
where $N$ is the number of samples. $z_i=f_e(h_i,\epsilon)$ is the extracted planning-relevant representation. $f_e$ is the information bottleneck encoder that maps the intermediate feature map $h_i$ to $z_i$. $\epsilon$ is a Gaussian random variable used for reparameterization. $r(Z)$ is a variational approximation to $p(Z)$ and here we set $r(Z)$ as a fixed Gaussian distribution following \cite{alemi2016deep}. $q_d(y_i^j|z_i)$ is a variational approximation to $p(y_i^j|z_i)$. $q_d$ is the information bottleneck decoder mapping $z_i$ to each planning  state $y_i^j$. 
The architectures of the information bottleneck encoder and decoder are described in Appendix A.

Rather than directly optimizing Eq.(\ref{ibobj}), we maximize its lower bound $\mathcal{L}_{IB}$ to effectively learn the planning-relevant feature $z$. After that,  we use $L_1$ loss to make the student model's planning-relevant feature $z^S$ match the teacher model's planning-relevant feature $z^T$:
\begin{equation}
\setlength{\abovedisplayskip}{0.02in}
\setlength{\belowdisplayskip}{0.02in}
    \mathcal{L}_{z} = \frac{1}{N} \sum_{i=1}^N |z^T_i - z^S_i|,    
\end{equation}

By minimizing $\mathcal{L}_{z}$, the student can inherit only planning-relevant knowledge from the intermediate layer of the teacher, instead of mimicking everything blindly. 

\vspace{-0.02in}
\subsection{Safety-aware Waypoint-attentive Distillation}
Within a planned trajectory, each waypoint holds varying importance for the motion planning task. Consequently, it is essential for the student model to prioritize the imitation of crucial waypoints generated by the larger teacher model.
To achieve this, we devise a safety-aware waypoint attention mechanism for distilling knowledge of waypoints.

\textbf{Waypoint attention weight.} Considering that the importance of each waypoint is related to the context of the driving scene, we determine the significance of each waypoint  by calculating  the attention weight between the BEV scene image $B \in \mathbb{R}^{C\times H\times W}$ and each waypoint $w_i$ in a trajectory $\mathcal{T} = \{w_i\in \mathbb{R}^{2}\}_{i=1}^T$. In order to incorporate position information into the BEV representation, we append the coordinates of each pixel to its channel dimension, resulting in $\tilde{B} \in \mathbb{R}^{(C+2)\times H\times W}$. The attention weight can be then calculated as follows:
\begin{equation}
\setlength{\abovedisplayskip}{0.05in}
\setlength{\belowdisplayskip}{0.05in}
    Q = f_{bev} (\tilde{B}), K = f_w (\mathcal{T}), A = softmax(\frac{QK}{\sqrt{d_k}}), 
\end{equation}
where $A = \{a_i\}_{i=1}^T$ and $a_i$ is the attention weight for each waypoint $w_i$. $f_{bev}$ and $f_w$ are the BEV encoder and waypoint encoder, respectively.
$d_k$ is the dimension of $K$. By doing so, the importance of each waypoint can be determined by incorporating its  contextual information from its driving environment.
The architectures of $f_{bev}$ and $f_w$ are described in Appendix A.

\textbf{Safety-aware ranking loss.} To promote the attention weight's awareness of safety-critical circumstances, we design a safety-aware ranking loss. First, we define a safety-aware kernel function $\psi_i$ as: 
\begin{equation}
\setlength{\abovedisplayskip}{0.05in}
\setlength{\belowdisplayskip}{0.05in}
    \psi_i = \sum_j \kappa_{ij} = \sum_j e^{-\frac{1}{2\sigma^2}{||p_i-p_j||}^2}, 
\end{equation}
where $\kappa_{ij}=e^{-\frac{1}{2\sigma^2}{||p_i-p_j||}^2}$ is a Gaussian kernel function \cite{zhong2013optimizing} that measures the proximity of the $i^{th}$ waypoint $w_i$ to $j^{th}$ moving obstacles. $p_i$ and $p_j$ are the positions of waypoint $w_i$ and $j^{th}$ moving obstacles, respectively. $\sigma$ is a hyper-parameter that adjusts the smoothness of the kernel function. 
By summing up $\kappa_{ij}$, the safety-aware kernel function $\psi_i$ 
 can effectively estimate the proximity of waypoint $w_i$ to other moving obstacles.
 
Intuitively, a large value of $\psi_i$ indicates that the significant and safe-critical nature of waypoint $w_i$, where a small deviation from its intended path could potentially lead to a collision with nearby moving obstacles.
 Hence, we design a pair-wise ranking loss to encourage  waypoints with larger values of $\psi_i$ to receive correspondingly greater attention weights:
\begin{equation}
\setlength{\abovedisplayskip}{0.02in}
\setlength{\belowdisplayskip}{0.02in}
    \mathcal{L}_{rank} = \sum_{i=1}^T \sum_{j=1}^T max(0, -r_{ij}(a_i-a_j)),
\end{equation}
where the comparison indicator $r_{ij}$ is defined as: $r_{ij} = 1$ if $\psi_i > \psi_j$, and $r_{ij} = -1$ otherwise. $a_i$ is the obtained attention weight for each waypoint $w_i$.
By minimizing $\mathcal{L}_{rank}$, the attention weight can effectively  determine the importance of each waypoint by taking safety into consideration. 

 \begin{table*}
\centering
\caption{Overall performance of motion planners of different size, with and without utilizing PlanKD, on the Town05 Long Benchmark. The inference time per frame is evaluated on GeForce RTX 3090 GPU. }
\vspace{-0.05in}
 \label{overall_long}
  \setlength{\tabcolsep}{1.5mm}
 {
\begin{tabular}{@{}c|ccc|ccccc@{}}
\toprule
Backbone          & \makecell[c]{Param\\Count} & \makecell[c]{Inference\\Time (ms)} & \makecell[c]{With\\PlanKD} & \makecell[c]{Driving\\Score($\uparrow$)} & \makecell[c]{Route\\Completion($\uparrow$)} & \makecell[c]{Infraction\\Score($\uparrow$)} & \makecell[c]{Collision\\Rate($\downarrow$)} & \makecell[c]{Infraction\\Rate($\downarrow$)} \\ 
\midrule
\multirow{7}{*}{InterFuser}    & 52.9M & 78.3 &- & 53.44 & 97.52 & 0.549  & 0.090  & 0.078      \\ \cmidrule(l){2-9} 
          & 26.3M & 39.7  &\ding{55} & 36.55 & 94.00    & 0.425 & 0.121 & 0.068  \\
 & 26.3M & 39.7 &\checkmark  & 55.90  & 97.44 & 0.562  & 0.094 & 0.093  \\ \cmidrule(l){2-9} 
        & 11.7M & 22.8 &\ding{55}  & 17.12 & 66.19 & 0.358  & 0.362 & 0.283 \\
 & 11.7M & 22.8 &\checkmark  & 28.79 & 80.50  & 0.430   & 0.315 & 0.202  \\ \cmidrule(l){2-9} 
           & 3.8M & 17.2 &\ding{55}  & 11.96 & 64.56 & 0.335  & 1.117 & 0.722  \\
  & 3.8M & 17.2 &\checkmark  & 26.15 & 70.95 & 0.410   & 0.361 & 0.265  \\ \midrule
\multirow{7}{*}{TCP}       & 25.8M & 17.9  &- & 53.41 & 100.0   & 0.534  & 0.076 & 0.115      \\ \cmidrule(l){2-9} 
  & 13.9M & 10.7 &\ding{55} & 39.96 & 91.06 & 0.443  & 0.183 & 0.157 \\ 
  & 13.9M & 10.7 &\checkmark & 53.19 & 93.28   & 0.579 & 0.084 & 0.116 \\ \cmidrule(l){2-9} 
  & 7.6M & 8.5 &\ding{55} & 25.88 & 52.69 & 0.690   & 0.110  & 0.101  \\
  & 7.6M & 8.5  &\checkmark & 35.44 & 63.95 & 0.673  & 0.096 & 0.087 \\ \cmidrule(l){2-9} 
  & 3.1M & 7.2 &\ding{55} & 16.16 & 31.33 & 0.781  & 0.098 & 0.161  \\
  & 3.1M & 7.2 &\checkmark & 23.84 & 32.03 & 0.858  & 0.052 & 0.074  \\ \bottomrule
\end{tabular}
}
\vspace{-0.1in}
\end{table*}

\textbf{Waypoint-attentive distillation.} 
After obtaining the attention weight, we incorporate it into the loss function for waypoint imitation as follows:
\begin{equation}
\setlength{\abovedisplayskip}{0.02in}
\setlength{\belowdisplayskip}{0.02in}
    \mathcal{L}_w = \sum_{i=1}^T a_i |w_i^S - w_i^T|, 
\end{equation}
where $\mathcal{L}_w$ is the safety-aware waypoint-attentive loss function.
$w_i^S, w_i^T$ are the waypoints of the student planner and the teacher planner, respectively. In addition, to avoid the student model becoming overly fixated on important waypoints at the expense of neglecting other waypoints, we introduce an entropy loss to ensure a smoother attention weight distribution by $\mathcal{L}_e = \sum_{i=1}^T a_i \log (a_i)$. 

\vspace{-0.03in}

\subsection{Optimization}

Our framework can be trained in an end-to-end fashion, and the overall loss function is defined as:
\begin{equation}
\setlength{\abovedisplayskip}{0.05in}
\setlength{\belowdisplayskip}{0.05in}
    \mathcal{L} = \mathcal{L}_w + \mathcal{L}_{w^*} - \mathcal{L}_{IB} + \alpha_z \mathcal{L}_z + \alpha_r \mathcal{L}_{rank} + \alpha_e \mathcal{L}_e,
\end{equation}
where $\alpha_z, \alpha_r, \alpha_e$ are hyper-parameters. Note that $\mathcal{L}_{w^*}$ is the $L_1$ loss of waypoints used to align with the expert trajectory. It serves as a source of ground-truth information and is weighted by the safety-aware attention, similar to $\mathcal{L}_w$.
$\mathcal{L}_{IB}$ represents the lower bound of the information bottleneck objective, which is expected to be maximized for updating the IB encoder and IB decoder.
By minimizing $\mathcal{L}$, the student planner could distill effective knowledge of motion planning  from both the perception and the motion producer modules of the teacher planner.
 The pseudo-code of training could be found in Appendix B.

%% file: sec/4_experiment.tex
\section{Experiments}

\subsection{Experimental Setup}
In this section, we introduce our experimental settings. 

\textbf{Evaluation task.}
We implement and evaluate our PlanKD for motion planning using version 0.9.10.1 of the CARLA simulator \cite{dosovitskiy2017carla}.  CARLA is widely recognized for simulating realistic driving scenarios. 
The task of the motion planner is to drive a vehicle towards a predefined destination, following a given route using high-level navigation commands.

\textbf{Datasets.}
We collect 800K frame data at 2 FPS from 8 public towns and 21 weather conditions offered by CALAR simulator,  similar to \cite{prakash2021multi, wutrajectory, shao2023safety}. Following \cite{prakash2021multi,hu2022st}, we use 7 towns for training and hold out Town05 for evaluation, due to its large diversity of driving scenarios and high densities of dynamic agents. We conduct evaluation on two Town05 benchmarks \cite{prakash2021multi}: Town05 Short Benchmark and Town05 Long Benchmark. The former includes 10 short routes ranging from 100-500m in length, each containing 3 intersections. The latter consists of 10 long routes spanning 1000-2000m, and each route includes 10 intersections.

\textbf{Evaluation metrics.}
We utilize three popular metrics in motion planning to evaluate our method: Driving Score (main metric), Route Completion and Infraction Score \cite{wutrajectory}. 
The Driving Score is defined as the product of Route Completion and Infraction Score. 
The Route Completion is the percentage of the route completed by the planner. 
The Infraction Score is a performance penalty factor that is initially set to 1.0. It gradually decreases by a certain percentage if the planner commits specific infractions, such as running a red light or colliding with pedestrians.

Besides, to intuitively evaluate the safety of the planners, we additionally defined two metrics: Collision Rate (\#/km) and Infraction Rate (\#/km). The Collision Rate represents the total number of collisions with pedestrians, vehicles, and environmental layout elements per kilometer traveled. The Infraction Rate quantifies the total number of infractions per kilometer, including instances of running red lights, disregarding stop signs, and driving off-road.

\textbf{Backbones and baselines.}
To demonstrate the versatility of our PlanKD, we apply it to compress two cutting-edge motion planning models, InterFuser \cite{shao2023safety} and TCP \cite{wutrajectory}. This showcases the seamless compatibility of PlanKD with different motion planners. Both of these models have achieved top rankings in the CARLA leaderboard  \cite{carlaleaderboard}. 
 For baselines, we utilize six lightweight planners by reducing the number of parameters of InterFuser and TCP respectively. 
The original-size InterFuser and TCP serve as the teacher model for the corresponding lightweight planners, respectively.
To further verify the effectiveness of our proposed PlanKD, we compare it with three typical knowledge distillation methods for model compression: AT \cite{zagoruyko2016paying}, ReviewKD \cite{chen2021distilling} and DPK \cite{zongbetter}. 
Please refer to Appendix A for the structures of the lightweight models and the implementation details.

\begin{table*}[]
\centering
\vspace{-0.1in}
\caption{Overall performance of motion planners of different size, with and without utilizing PlanKD, on the Town05 Short Benchmark. The inference time per frame is evaluated on GeForce RTX 3090 GPU.}
\vspace{-0.05in}
 \label{overall_short}
  \setlength{\tabcolsep}{1.5mm}
 {
\begin{tabular}{@{}c|ccc|ccccc@{}}
\toprule
Backbone      & \makecell[c]{Param\\Count} & \makecell[c]{Inference\\Time (ms)} & \makecell[c]{With\\PlanKD}   & \makecell[c]{Driving\\Score($\uparrow$)} & \makecell[c]{Route\\Completion($\uparrow$)} & \makecell[c]{Infraction\\Score($\uparrow$)} & \makecell[c]{Collision\\Rate($\downarrow$)} & \makecell[c]{Infraction\\Rate($\downarrow$)} \\ 
\midrule
\multirow{7}{*}{InterFuser}                      & 52.9M & 78.3  &- & 94.88         & 99.88            & 0.950             & 0.141          & 0.000                            \\ \cmidrule(l){2-9} 
            & 26.3M & 39.7   &\ding{55} & 82.70          & 96.87            & 0.841           & 0.662          & 0.141                   \\
 & 26.3M & 39.7  &\checkmark   & 93.69         & 96.43            & 0.960             & 0.207          & 0.000                       \\ \cmidrule(l){2-9} 
            & 11.7M & 22.8  &\ding{55}  & 58.17         & 89.69            & 0.644           & 0.731          & 1.638                   \\
& 11.7M & 22.8  &\checkmark  & 74.57         & 96.65            & 0.762            & 0.479          & 0.622                    \\ \cmidrule(l){2-9} 
              & 3.8M  & 17.2   &\ding{55} & 54.56         & 79.11            & 0.688           & 0.503          & 1.603                     \\
 & 3.8M  & 17.2  &\checkmark  & 65.18         & 81.52            & 0.807            & 0.337          & 0.863             \\ \midrule
 
\multirow{7}{*}{TCP}                      & 25.8M   & 17.9 &-  & 95.53 & 99.53 & 0.960   & 0.141 & 0.000         \\ \cmidrule(l){2-9} 
              & 13.9M & 10.7 &\ding{55}  & 84.53 & 88.53 & 0.960   & 0.141 & 0.000     \\
 & 13.9M & 10.7  &\checkmark  & 95.49 & 99.49 & 0.960 & 0.141 & 0.000      \\ \cmidrule(l){2-9} 
               & 7.6M & 8.5  &\ding{55} & 79.34 & 86.84 & 0.919  & 0.303 & 0.000     \\
 & 7.6M & 8.5  &\checkmark  & 83.59 & 87.09 & 0.960   & 0.162 & 0.000     \\ \cmidrule(l){2-9} 
              & 3.1M & 7.2     &\ding{55} & 58.70  & 65.17 & 0.930   & 0.162 & 0.143 \\
 & 3.1M & 7.2   &\checkmark & 64.86 & 68.36 & 0.960   & 0.162 & 0.000     \\ \bottomrule
\end{tabular}
}
\end{table*}

\begin{table*}[]
\centering
\vspace{-0.05in}
\caption{Comparison with other knowledge distillation methods on Town05 Short Benchmark.}
\vspace{-0.05in}
 \label{comparison_short}
  \setlength{\tabcolsep}{1mm}
 {
\begin{tabular}{@{}ccccccccc@{}}
\toprule
Method       & \makecell[c]{Backbone}   & \makecell[c]{Teacher\\Param} & \makecell[c]{Student\\Param}  & \makecell[c]{Driving\\Score($\uparrow$)} & \makecell[c]{Route\\Completion($\uparrow$)} & \makecell[c]{Infraction\\Score($\uparrow$)} & \makecell[c]{Collision\\Rate($\downarrow$)} & \makecell[c]{Infraction\\Rate($\downarrow$)} \\ \midrule
AT           &\multirow{4}{*}{InterFuser}  & 52.9M & 26.3M & 83.94 & 99.88 & 0.839 & 0.426 & 0.141 \\
ReviewKD      & & 52.9M & 26.3M & 84.19 & 99.49 & 0.845 & 0.569 & 0.000     \\
DPK           & & 52.9M & 26.3M & 84.59 & 99.49 & 0.850 & 0.286 & 0.283 \\
PlanKD (Ours) & & 52.9M & 26.3M & 93.69 & 96.43 & 0.960   & 0.207 & 0.000     \\ \midrule
AT          &\multirow{4}{*}{TCP}   & 25.8M & 13.9M & 86.66  & 89.37  & 0.960  & 0.209  & 0.000     \\
ReviewKD     & & 25.8M & 13.9M & 88.46 & 95.17 & 0.930   & 0.149 & 0.145 \\
DPK          & & 25.8M & 13.9M & 88.54 & 92.34 & 0.960 & 0.149 & 0.000       \\
PlanKD (Ours) & & 25.8M & 13.9M & 95.49 & 99.49 & 0.960 & 0.141 & 0.000     \\\bottomrule
\end{tabular}
}
\vspace{-0.2in}
\end{table*}

\subsection{Overall Performance}
In this section, we evaluate the overall performance of our PlanKD, as shown in Table \ref{overall_long} and \ref{overall_short}.
The symbol \ding{55} in the column `With PlanKD' represents that we directly train the models by imitating the expert, while the symbol \checkmark means that we adopt PlanKD to train the model.
If without PlanKD, the performance of TCP and InterFuser significantly drops as the number of parameters decreases. 
However, by employing PlanKD, we observe a significant improvement in the performance of these lightweight planners. 
Remarkably, when the number of parameters is halved, the planning models trained with PlanKD remains comparable or even better performance than that of the original models, while lowering the reference time by approximately 50\%.
Furthermore, PlanKD generate a safer lightweight motion planner, exhibiting significantly lower collision rates and infraction rates compared to models of the same size without PlanKD.
These findings demonstrate PlanKD can serve as a portable and safe solution for resource-limited deployment.

\begin{table*}[]
\centering
\caption{Ablation Study of PlanKD on Town05 Short Benchmark.}
\vspace{-0.05in}
 \label{ablation_short}
  \setlength{\tabcolsep}{1mm}
 {
\begin{tabular}{@{}ccccccccc@{}}
\toprule
Method     & \makecell[c]{Backbone}       & \makecell[c]{Teacher\\Param} & \makecell[c]{Student\\Param}  & \makecell[c]{Driving\\Score($\uparrow$)} & \makecell[c]{Route\\Completion($\uparrow$)} & \makecell[c]{Infraction\\Score($\uparrow$)} & \makecell[c]{Collision\\Rate($\downarrow$)} & \makecell[c]{Infraction\\Rate($\downarrow$)} \\ \midrule
PlanKD-w.o.-entropy        &\multirow{4}{*}{InterFuser}  & 52.9M & 26.3M  & 84.00      & 100.0       & 0.839        & 0.424       & 0.141      \\
PlanKD-w.o.-safe-att & & 52.9M & 26.3M  & 85.66 & 96.83 & 0.869 & 0.555 & 0.141     \\
PlanKD-w.o.-IB             & & 52.9M & 26.3M& 88.49 & 99.49 & 0.890   & 0.283 & 0.141 \\
PlanKD                      & & 52.9M & 26.3M & 93.69 & 96.43 & 0.960   & 0.207 & 0.000     \\ \midrule
PlanKD-w.o.-entropy       &\multirow{4}{*}{TCP}   & 25.8M & 13.9M       & 78.59      & 94.67      & 0.839       & 0.566    & 0.141      \\
PlanKD-w.o.-safe-att & & 25.8M & 13.9M  & 88.83 & 92.62 & 0.960  & 0.149 & 0.000 \\
PlanKD-w.o.-IB             & & 25.8M & 13.9M  & 91.32 & 95.11 & 0.960   & 0.149 & 0.000 \\
PlanKD                     & & 25.8M & 13.9M  & 95.49 & 99.49 & 0.960 & 0.141 & 0.000 \\ \bottomrule
\end{tabular}
}
\end{table*}

\begin{figure*}
\vspace{-0.05in}
\centering
\subcaptionbox{}{\includegraphics[width = .195\linewidth]{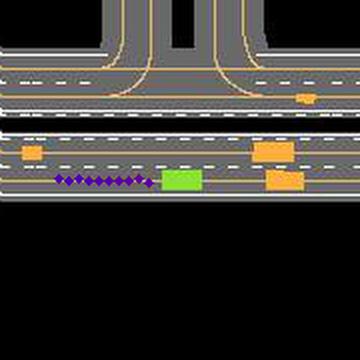}}
\subcaptionbox{}{\includegraphics[width = .195\linewidth]{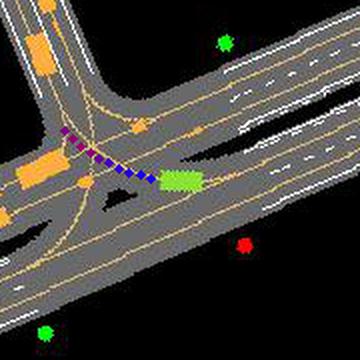}}
\subcaptionbox{}{\includegraphics[width = .195\linewidth]{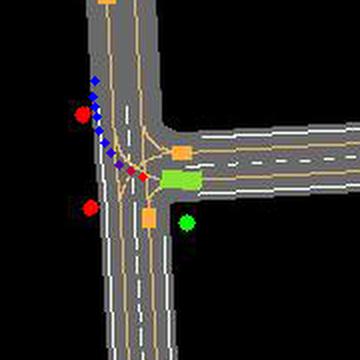}}
\subcaptionbox{}{\includegraphics[width = .195\linewidth]{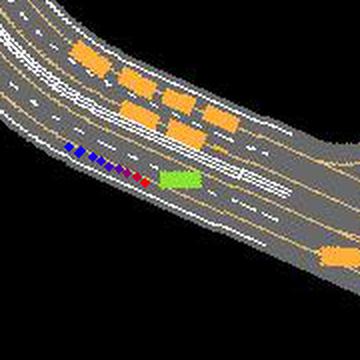}}
\subcaptionbox{}{\includegraphics[width = .195\linewidth]{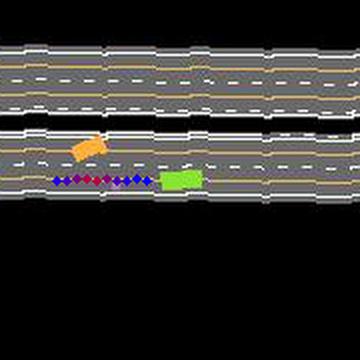}}
\vspace{-0.1in}
\caption{Visualizations of safety-aware attention weights under different driving scenarios. The green block denotes the ego-vehicle and the yellow blocks represent other road users (e.g. vehicles, bicycles). The redder a waypoint is, the higher attention weight it has.}
\vspace{-0.15in}
\label{visual}
\end{figure*}

\subsection{Comparison with Knowledge Distillation}
To further evaluate our PlanKD, we compare it with three popular knowledge distillation methods, i.e., AT \cite{zagoruyko2016paying}, ReviewKD \cite{chen2021distilling}, DPK \cite{zongbetter}. As shown in Table \ref{comparison_short}, our PlanKD outperforms these methods by a large margin. The reasons are as follows: First, existing knowledge distillation methods don't focus on distilling the knowledge that are significant to planning to the student, which could cause invalid knowledge transferring; Second, they fail to take the safety of the small planners into account, resulting in a large collision rate.
Besides, one interesting finding is that the Route Completion of PlanKD is sometimes lower than other KD methods, this might because the model trained by our PlanKD prioritizes safety by making the decision to stop when encountering poor traffic conditions.
In summary, PlanKD is a superior knowledge distillation method for compressing the motion planner. 

\subsection{Ablation Study}
We perform ablation study to examine the effectiveness of each component in our method. Specifically, we design three variants of PlanKD: PlanKD-w.o.-entropy represents our method without using the entropy loss in the safety-aware attention; PlanKD-w.o.-safe-att denotes our method without using the safety-aware attention in the waypoint distillation; PlanKD-w.o.-IB stands for our method distilling the whole feature map instead of the planning-relevant feature in the intermediate layer. 
As shown on Table \ref{ablation_short}, PlanKD outperforms PlanKD-w.o.-entropy. This illustrates that only focusing on important waypoints and totally neglecting other waypoints could harm the performance.
Besides, PlanKD obtains better performance than PlanKD-w.o.-safe-att, demonstrating that our safety-aware attention mechanism is able to determine the importance of each waypoints for distillation.
Finally, PlanKD has superiority over PlanKD-w.o.-IB, illustrating learning planning-relevant feature to transfer is beneficial for motion planning. 

\subsection{Visualizations}
To intuitively show the effectiveness of the safety-aware attention mechanism, we visualize  the attention weights under different driving scenes. 
Figure \ref{visual} (a) describes a normal going straight scene without potential risks. In this case, the attention weight is uniform across all the waypoints.
Figure \ref{visual} (b) and Figure \ref{visual} (c) show the attention weights when the ego-vehicle pass an intersection. It is obvious that the waypoints near the interactions with other vehicles have larger attention weight because these waypoints are safety-critical. 
Figure \ref{visual} (d) depicts a lane-changing scenario where the model assigns larger weights to the first few waypoints due to the likelihood of potential interactions with other vehicles. 
These safety-critical waypoints require extra caution to ensure the safety of the ego-vehicle and other road users.
In Figure \ref{visual} (e), the ego-vehicle encounters a vehicle cutting into its lane, and the waypoints near the merging point have larger attention weights. This is because these waypoints are crucial for avoiding collisions.
Overall, our method can effectively assign larger attention weights to safety-critical waypoints  during distillation, ensuring the safety of the student motion planner. Futhermore, we also visualize the intermediate feature maps to investigate the planning-relevant knowledge extracted by our method in Appendx C.

%% file: sec/5_conclusion.tex
\section{Conclusion}
In this paper, we propose PlanKD, a knowledge distillation method tailored for compressing the end-to-end motion planner. The proposed method can learn  planning-relevant features via information bottleneck for effective feature distillation. Moreover, we design a safety-aware waypoint-attentive distillation mechanism to adaptively decide the significance of each waypoint for waypoint distillation. Extensive experiments verify the effectiveness of our method, demonstrating PlanKD can serve as a portable and safe solution for resource-limited deployment. 

\section{Acknowledgment}
This work was supported by the NSFC under Grants 62122013, U2001211. This work was also supported by the Innovative Development Joint Fund Key Projects of Shandong NSF under Grants ZR2022LZH007.

%% file: sec/X_suppl.tex
\clearpage
\setcounter{page}{1}
\maketitlesupplementary

\appendix
\noindent\textbf{\Large{Appendix}}


\vspace{0.1in}

\section{Implementation Details}

\begin{table*}
\centering
\caption{Configurations of different planners. Transformer-3 (128) denotes a 3-layer transformer with an embedding size of 128. MLPs-half denotes MLPs with half of the original hidden size. The inference time per frame is evaluated on GeForce RTX 3090 GPU.}
 \label{config}
  \setlength{\tabcolsep}{1.2mm}
 {
\begin{tabular}{@{}ccccccc@{}} \toprule
Backbone       & \makecell[c]{Parameter\\Count}     & \makecell[c]{Camera Perception\\Backbone} & \makecell[c]{LiDAR Perception\\Backbone} & \makecell[c]{Motion Producer\\Backbone}  & \makecell[c]{Model\\FLOPS}  & \makecell[c]{Inference\\Time (ms)} \\ \midrule
\multirow{4}{*}{InterFuser}   & 52.9M    & ResNet-50                  & ResNet-18                 & Transformer-6 (256)          & 46.51G & 78.3     \\
 & 26.3M   & ResNet-18                 & ResNet-18                 & Transformer-3 (128)        & 25.52G & 39.7       \\
 & 11.7M  & ResNet-10                  & ResNet-10                 & Transformer-3 (64)           & 11.12G  & 22.8        \\
 & 3.8M   & ResNet-6                   & ResNet-6                  & Transformer-2 (64)         & 7.21G  &  17.2        \\ \midrule
\multirow{4}{*}{TCP}          & 25.8M    & ResNet-34                  &-                           & MLPs                      & 17.09G &  17.9        \\
 & 13.9M         & ResNet-18                  &-                           & MLPs-half        & 8.47G &  10.7   \\
   & 7.6M      & ResNet-10                  &-                           & MLPs-half           & 4.15G &  8.5  \\
  & 3.1M      & ResNet-6                   &-                           & MLPs-half     & 	2.67G &   7.2 \\ \bottomrule
\end{tabular}
}
\end{table*}

\vspace{0.1in}

\subsection{Hyper-parameter Setting}
We perform all the experiments using GeForce RTX 3090 GPU. 
As for training, we use the Adam optimizer \cite{kingma2014adam} for optimization with a learning rate $0.005$ for InterFuser and $0.0001$ for TCP. 
For TCP backbones, the epoch number is set to 30 and the batch size is set to 24. For InterFuser backbones, the epoch number is set to 10 and the batch size is set to 16.
 We empirically set $\alpha_r=0.1$. $\alpha_z$ is set to $0.1$ for InterFuser and $0.5$ for TCP. $\alpha_e$ is set to $0.005$ for InterFuser and $0.05$ for TCP. The standard deviation $\sigma$ in the kernel function for adjusting the smoothness is set to $3$. The action threshold $\delta$ is set to 0.1. The Lagrange multiplier $\beta$ is set to 0.001. 
Following the original papers \cite{wutrajectory, shao2023safety}, we set the number of planned waypoints to $T=4$ for TCP and $T=10$ for InterFuser.

\vspace{0.1in}

\subsection{Image Resolution Setting}
For InterFuser,  the resolution of camera image is $224\times224$ for the front camera and $128\times128$ for the side camera, following the original paper \cite{shao2023safety}. For TCP, the resolution of camera image is $900\times256$ for the front camera,  following the original paper \cite{wutrajectory}. The horizontal field of view for all cameras is set as $100^{\circ}$. In our method, the BEV scene image is provided by the CARLA simulator \cite{dosovitskiy2017carla} and is with a resolution of $180\times180$. 

\vspace{0.1in}

\subsection{Lightweight Planner Architecture Setting} As aforementioned in the main body of the paper, the end to end motion planners can generally be divided into two main parts: the perception backbone and the motion producer. In this paper, we reduce the number of the parameters of these two parts by taking smaller backbones as lightweight planners. The detailed configurations of the original planners and the corresponding lightweight planners are presented in Table \ref{config}.

\vspace{0.1in}

\subsection{Planning-relevant Feature Distillation Module Setting}
In the planning-relevant feature distillation module, we configure both the encoder and decoder of the information bottleneck by using a 3-layer MLPs with a hidden size of 512 and LeakyReLU \cite{maas2013rectifier}  as the activation function. The dimension of the planning-relevant feature is set to 256.
To transfer planning-relevant knowledge, We empirically select the middlemost layer of the teacher's perception backbone to distill the knowledge to the middlemost layer of the the student's perception backbone. Before inputting the intermediate feature map to the information bottleneck encoder, we perform channel-wise averaging along the channel dimension.

\vspace{0.1in}

\subsection{Safety-aware Waypoint-attentive Distillation Module Setting}
In the safety-aware waypoint-attentive distillation module, the BEV encoder consists of a 6-layer CNN followed by a 2-layer MLP with a hidden size of 512. The waypoint encoder, on the other hand, is configured as a 2-layer MLP with a hidden size of 128. Both of these two encoders utilize the LeakyReLU activation function. For simplicity, we adopt the expert waypoints as the teacher waypoints in this paper. In the attention mechanism, both of the dimensions of the query and the key are set to 64.

\vspace{0.1in}

\section{Training Pseudo-code}
The training procedure for our PlanKD method is outlined in Algorithm \ref{training}. Firstly, the student planner and teacher planner undergo forward propagation to obtain their intermediate feature maps and output waypoints. These intermediate feature maps are then passed through the information bottleneck encoder to extract the planning-relevant features. Using the planning-relevant features from both the teacher planner and the student planner, we calculate the planning-relevant knowledge distillation loss. Next, the planning-relevant features are input to the information bottleneck decoder to obtain the planning states, which are used to compute the upper bound of the information bottleneck objective. Moving on to the safety-aware waypoint-attentive distillation module, we determine the importance of the teacher's waypoints and the expert's waypoints. Based on the obtained importance weights, we calculate the safety-aware waypoint loss, as well as the safety-aware ranking loss and the entropy loss. Finally, all these losses are aggregated and used as the overall loss for optimization. By employing PlanKD during the training process, we can develop a portable and safe planner suitable for deployment in resource-limited environments.

\begin{algorithm*}
  \setcounter{algorithm}{0}
  \caption{The training procedure of PlanKD}  
  \begin{algorithmic}[0]  
    \Require 
    a pretrained large teacher planner $\mathcal{F}_{\theta}^T$, dataset $\mathcal{D}=\{(I,\mathcal{T}^*)\}$, ground truth planning states $Y^i$, BEV scene representation $B$, epochs $N_e$;
    \Ensure
    a trained compact student planner $\mathcal{F}_{\phi}^S$;
    \State Initialize the parameters of student planner $\mathcal{F}_{\phi}^S$;
    \State Initialize the parameters of the two modules in PlanKD;
    \State Freeze the parameters of teacher planner $\mathcal{F}_{\theta}^T$;
    \For{each epoch $e$ from $1$ to $N_e$}
        \For{each batch $b$ in epoch $e$}
            \State obtain the intermediate feature map $h^T$ of teacher planner $\mathcal{F}_{\theta}^T$; 
            \State obtain the intermediate feature map $h^S$ of student planner $\mathcal{F}_{\phi}^S$;
            \State input $h^T$,$h^S$ to IB encoder to derive planning-relevant feature $z^T$,$z^S$; 
            \State calculate the planning-relevant knowledge distillation loss $\mathcal{L}_z$;
            \State input $z^T$,$z^S$ to IB decoder to derive the prediction of planning states; 
            \State calculate the upper bound of the information bottleneck objective $\mathcal{L}_{IB}$;
            \State derive the attention weight between teacher waypoint $w_i^T$ and $B$; 
            \State derive the attention weight between expert waypoint $w_i^*$ and $B$; 
            \State calculate the safety-aware waypoint loss $\mathcal{L}_w$ and $\mathcal{L}_{w^*}$;
            \State calculate the ranking loss $\mathcal{L}_{rank}$ the entropy loss $\mathcal{L}_e$;
            \State calculate the overall loss $\mathcal{L}$;
            \State optimize the learnable parameters by $\mathcal{L}$;
        \EndFor
    \EndFor

  \end{algorithmic}  
  \label{training}
\end{algorithm*}

\begin{table*}
\centering
\caption{Comparisons with other knowledge distillation methods on the Town05 Long Benchmark.}
 \label{comparison_long}
  \setlength{\tabcolsep}{1mm}
 {
\begin{tabular}{@{}ccccccccc@{}}
\toprule
Method     & \makecell[c]{Backbone}      & \makecell[c]{Teacher\\Param} & \makecell[c]{Student\\Param}  & \makecell[c]{Driving\\Score($\uparrow$)} & \makecell[c]{Route\\Completion($\uparrow$)} & \makecell[c]{Infraction\\Score($\uparrow$)} & \makecell[c]{Collision\\Rate($\downarrow$)} & \makecell[c]{Infraction\\Rate($\downarrow$)} \\ \midrule
AT            &\multirow{4}{*}{InterFuser}  & 52.9M & 26.3M   
& 41.62         & 85.61            & 0.472            & 0.112          & 0.134\\
ReviewKD      &  & 52.9M & 26.3M       & 40.67         & 93.25            & 0.426            & 0.178          & 0.168         \\
DPK            &  & 52.9M & 26.3M   & 44.29         & 81.10             & 0.550             & 0.095          & 0.113           \\
PlanKD (Ours)  &  & 52.9M & 26.3M   & 55.90          & 97.44            & 0.562            & 0.094          & 0.093           \\ \midrule
AT             &\multirow{4}{*}{TCP}   & 25.8M & 13.9M        & 43.31         & 100.0              & 0.433            & 0.159          & 0.128            \\
ReviewKD        &   & 25.8M & 13.9M         & 41.27         & 94.64            & 0.431            & 0.148          & 0.147           \\
DPK            &   & 25.8M & 13.9M      & 43.83 & 90.27 & 0.499 & 0.158 & 0.146             \\
PlanKD (Ours)  &   & 25.8M & 13.9M      & 53.19         & 93.28            & 0.579            & 0.084          & 0.116     \\ \bottomrule
\end{tabular}
}
\end{table*}

\begin{table*}
\centering
\vspace{0.1in}
\caption{Ablation Study of PlanKD on the Town05 Long Benchmark.}
 \label{ablation_long}
  \setlength{\tabcolsep}{1mm}
 {
\begin{tabular}{@{}ccccccccc@{}}
\toprule
Method        & \makecell[c]{Backbone}    & \makecell[c]{Teacher\\Param} & \makecell[c]{Student\\Param}  & \makecell[c]{Driving\\Score($\uparrow$)} & \makecell[c]{Route\\Completion($\uparrow$)} & \makecell[c]{Infraction\\Score($\uparrow$)} & \makecell[c]{Collision\\Rate($\downarrow$)} & \makecell[c]{Infraction\\Rate($\downarrow$)} \\ \midrule
PlanKD-w.o.-entropy       &\multirow{4}{*}{InterFuser}  & 52.9M & 26.3M    & 46.73         & 70.49            & 0.643            & 0.141          & 0.063           \\
PlanKD-w.o.-safe-att  &  & 52.9M & 26.3M & 44.55         & 75.37            & 0.555            & 0.141          & 0.097           \\
PlanKD-w.o.-IB              &  & 52.9M & 26.3M & 50.17         & 92.72            & 0.509            & 0.162          & 0.111           \\
PlanKD                       &  & 52.9M & 26.3M & 55.90          & 97.44            & 0.562            & 0.094          & 0.093           \\ \midrule
PlanKD-w.o.-entropy         &\multirow{4}{*}{TCP}   & 25.8M & 13.9M         & 45.72         & 71.64            & 0.668            & 0.088          & 0.127           \\
PlanKD-w.o.-safe-att  &   & 25.8M & 13.9M         & 45.07         & 100.0              & 0.450             & 0.160           & 0.121           \\
PlanKD-w.o.-IB             &   & 25.8M & 13.9M         & 50.70          & 100.0              & 0.507            & 0.096          & 0.130            \\
PlanKD                    &   & 25.8M & 13.9M         & 53.19         & 93.28            & 0.579            & 0.084          & 0.116       \\ \bottomrule
\end{tabular}
}
\end{table*}

\section{Additional Experiments}
\subsection{Additional Comparison with KD}
Here, we present additional comparison results with other knowledge distillation methods on the Town05 Long Benchmark. As shown in Table \ref{comparison_long}, it is evident that our PlanKD method continues to outperform previous knowledge distillation methods by a significant margin.

\subsection{Additional Ablation Study}
To further validate the effectiveness of our method, we perform an ablation study on the Town05 Long Benchmark, as presented in Table \ref{ablation_long}. The results further demonstrate the effectiveness of each component in our proposed method.

\subsection{Additional Visualizations}

\begin{figure*}
  \centering
  \includegraphics[width=0.9\linewidth]{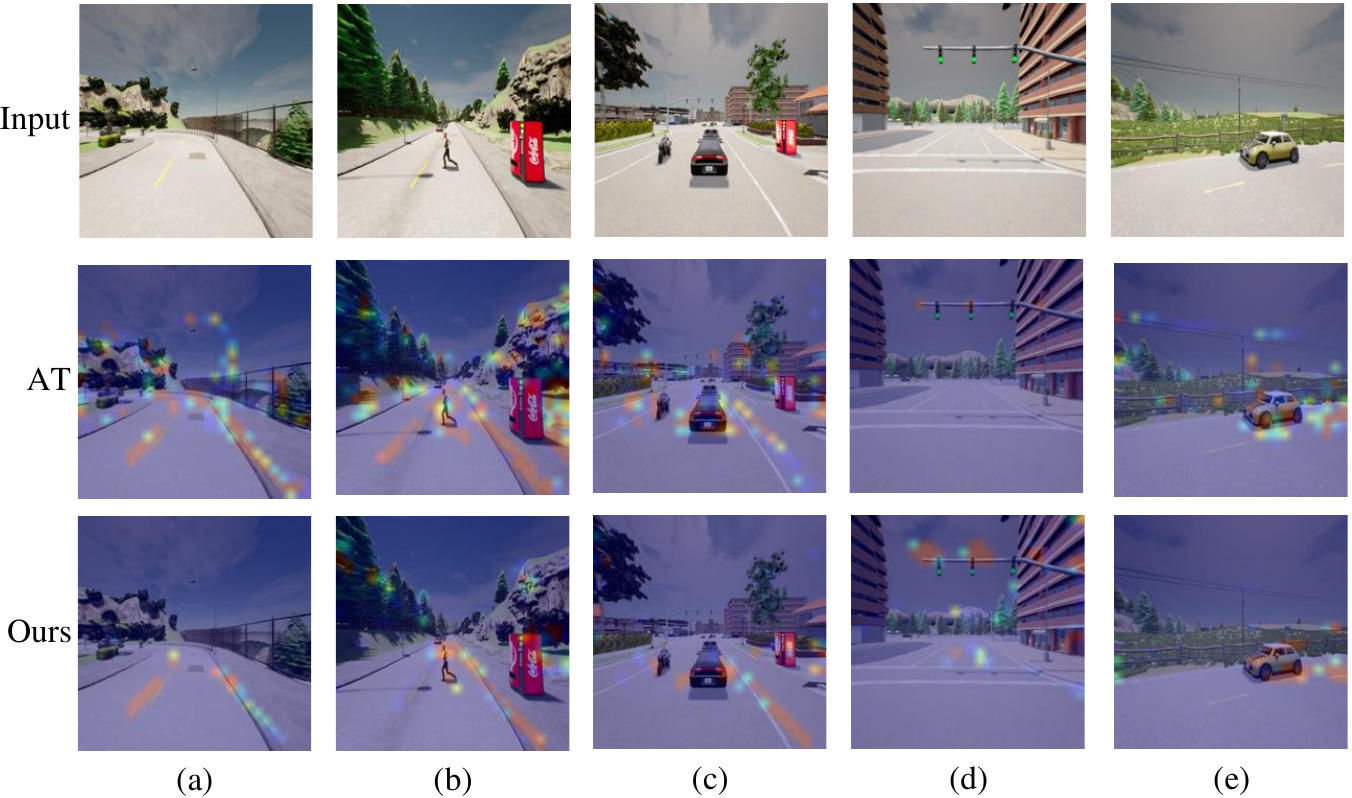}
  \caption{
     Visualizations of the intermediate feature maps of InterFuser. The redder regions represent higher activation values. The first row is the input image of the front camera. The second row is the corresponding attention map generated by AT \cite{zagoruyko2016paying}. The third row is the corresponding Grad-CAM \cite{selvaraju2017grad} visualization guided by the gradient of the planning states in the information bottleneck.
  }
  \label{vis}
\end{figure*}

To investigate the planning-relevant knowledge extracted by the information bottleneck, we employ the Grad-CAM technique \cite{selvaraju2017grad} to visualize the intermediate feature maps of InterFuser. The visualization is guided by the gradient of the planning states within the information bottleneck, revealing where the extracted planning-relevant knowledge is concentrated. The results are presented in Figure \ref{vis}. Figure \ref{vis}(a) represents a normal scene with no moving obstacles. The planning-relevant knowledge focuses on the lanes, indicating the importance of keeping lane for the ego-vehicle. In Figure \ref{vis}(b), where a pedestrian suddenly appears, the planning-relevant knowledge is directed towards the pedestrian, highlighting the need to avoid collision. Figure \ref{vis}(c) showcases a situation where a vehicle is in front and a motorbike is driving towards the ego-vehicle. In this case, it's important to maintain a safe distance, thus the planning-relevant knowledge emphasizes other road users. Figure \ref{vis}(d) depicts a scenario with a traffic light, where the attention is drawn to the state of the traffic light. Finally, Figure \ref{vis}(e) shows an intersection scenario where the ego-vehicle requires extra caution to interact with other road users. Thus,  the planning-relevant knowledge focuses on the interacting vehicle in front.
These visualizations indicate that our method can successfully extracts the  knowledge that are significant to planning across various scenarios.

Besides, we also visualize the attention maps generated by the knowledge distillation method AT \cite{zagoruyko2016paying}.  It can be observed that the generated attention maps contain numerous planning-irrelevant information (especially in Figure \ref{vis}(a)(b)(c)). This further indicates the superiority of our method.


\section{Limitations and Future Works}
Our work mainly focus on the knowledge distillation technique for compressing end-to-end motion planner in autonomous driving. 
Exploring the integration of other model compression techniques, such as quantization and pruning, into our approach is a promising avenue for future research. By doing so, we can further reduce the size of the motion planner and enhance its efficiency.

Besides, we devise a simple yet effective way to take the safety significance of each waypoint into account via the learning-based attention. In the future, it is possible to incorporate specific expert knowledge about driving to design a more comprehensive and refined strategy for determining the importance of waypoints.  In addition,  the current method primarily emphasizes the proximity of waypoints to obstacles as a measure of danger, which captures an important aspect of safety. The approach is grounded in the fact that immediate physical distance from obstacles is a critical factor in potential collisions. While our current approach prioritizes spatial proximity to obstacles, incorporating temporal aspects, could indeed offer a more comprehensive safety assessment.

Furthermore, in our approach, knowledge transfer in the intermediate layer is currently limited to feature maps within the same sensor modality. For planners that incorporate multiple sensor modalities, a potential future direction could involve developing methods to distill knowledge between different sensors to facilitate cross-modal knowledge transfer.

Finally,  our method trained on CARLA is subject to the well-known simulation-to-reality gap, which implies that its performance might differ when deployed in the real world. This necessitates extensive real-world testing and validation to ensure that the model's behavior aligns with expected safety norms. Safety assurance processes must encompass a wide range of scenarios and edge cases that vehicles might encounter, ensuring the model's robustness and reliability.